\documentclass[twoside,11pt]{article}

\usepackage{jmlr2e}

\usepackage{algorithm}
\usepackage{algpseudocode}
\usepackage{amsmath}
\usepackage{graphicx}
\usepackage{bbm}
\usepackage{tabulary}
\usepackage{booktabs}
\usepackage[symbol]{footmisc}
\usepackage{tikz}
\usepackage[frozencache=true,cachedir=.]{minted}
\usepackage{float}

\jmlrheading{1}{2021}{1-48}{4/00}{10/00}{Liu \& Luo}

\ShortHeadings{Monte Carlo Tree Search for the JVM}{Liu \& Luo 2021}
\firstpageno{1}

\begin{document}

\author{\name Larkin Liu \email larkin.liu@mail.utoronto.ca \\
       \addr University of Toronto \\
       \\
       \name Jun Tao Luo \email jtluo@andrew.cmu.edu \\
       \addr Carnegie Mellon University \\
}

\title{An Extensible and Modular Design and Implementation of Monte Carlo Tree Search for the JVM}

\maketitle

\begin{abstract}
Flexible implementations of Monte Carlo Tree Search (MCTS), combined with domain specific knowledge and hybridization with other search algorithms, can be very powerful for finding the solutions to problems in complex planning. We introduce \textit{mctreesearch4j}, a MCTS implementation written as a standard JVM library following key design principles of object oriented programming. We define key class abstractions allowing the MCTS library to flexibly adapt to any well defined Markov Decision Process or turn-based adversarial game. Furthermore, our library is designed to be modular and extensible, utilizing class inheritance and generic typing to standardize custom algorithm definitions. We demonstrate that the design of the MCTS implementation provides ease of adaptation for unique heuristics and customization across varying Markov Decision Process (MDP) domains. In addition, the implementation is reasonably performant and accurate for standard MDP's. Furthermore, the nuances of different types of MCTS algorithms are discussed. \\
\end{abstract}


\section{Introduction}

Sampling based strategies for the optimal planning of Markov Decision Processes (MDP) and adversarial games have been of great importance to the improvement to discovery of approximate solutions for complex high dimensionality domains where a full solution using complete methods are often infeasible. Algorithms such as Monte Carlo Tree Search (MCTS) \citep{Kocsis:2006} have displayed a promising ability to generate competitive solutions for domains exhibiting high dimensionality search spaces. 

In recent memory, MCTS used in combination with other game solving strategies have shown to be the world's state-of-the-art for high branching factor games such as Go \citep{Silver:2016}. Other algorithms also frequently apply MCTS to solve for games with high branching factors such as Kriegspiel \citep{Ciancarini:2010} and Othello \citep{Norvig:1992}. Our objective in this software is to create a extensible implementation of MCTS which is adaptable into multiple domains, applicable to either single player of multi player games.

\subsection{Discrete Markov Decision Process} \label{sec:mcts_intro}

The Markov Decision Process (MDP), described in \citep{Puterman:1994}, is a well-developed framework for planning under uncertainty. The application of MDP's can be found in various optimization strategies to prescribe an optimal set of actions given a stochastic environment where perfect observations are available. An $MDP\langle S, A, \mathbb{P}, R\rangle$ designates a set of states $S$, where the agent traverses from $S_t$ to $S_{t+1}$, for a horizon of $T$ in $T$ distinct time increments $t$ (starting from $t = 1$). $A$ is the action space, which is the set of actions the agent can take at any $t$ and may or may not be a function of $t$ itself. $\mathbb{P}$ is the probability space that outlines the transition probability space for state-to-state transitions for any action taken at time $t$, $\mathbb{P} : \{ S_{t} \times A \times S_{t+1} \} \to \{R \in \mathbb{R}\}$ subsequently returning a reward value, which belongs to $\mathbb{R}$. 

In a fundamental MDP, the state at any time $S_t$ is visible to the agent, and a corresponding reward is associated with each state. Therefore, if the state is observable, the value of each observable state is the reward associated with the state, and the availability of actions leading to rewards in the future. Evidently, they can form a recursive pattern to determine the value of the state, which can be solved via dynamic programming. Suppose we use $Q(S_t, a)$ to denote the immediate value when accounting for the immediate reward as well as the transitioned next state $S_{t+1}$. Where the Value of any state is is denoted by $V(s)$. 

\begin{equation} \label{mdp:q}
Q(S_t, a) = R(S_t, a) + \gamma \sum_{S_{t+1} \in S } P(S_{t+1}| S_t, a) V(S_{t+1})
\end{equation}

\begin{equation} \label{mdp:value}
V(S) = \operatorname*{argmax}_{a \in A}Q(S, a)
\end{equation}

Traditionally the exact solution to Eq. (\ref{mdp:value}) can be found using dynamic programming algorithms such as Value Iteration or Policy Iteration \citep{Bellman:1957}. Nevertheless, if the state space or branching factor is so large that traditional solutions such as dynamic programming cannot produce a solution within reasonable time or resource constraints, approximate methods via MCTS is a viable alternative for producing approximate optimal policies. The optimal policy denoted $\pi(S, a)$, is therefore the action, $a$ that maximizes Eq. (\ref{mdp:value}) at any particular state. 

\subsection{Monte Carlo Tree Search Algorithm} \label{sec:mcts_intro}

Monte Carlo Tree Search (MCTS) serves as an approach to solve the optimal policy for any well-defined MDP that does not always involve an exhaustive search through the entire search space. It has the potential to provide feasible solutions to high dimension and high branching MDP's, where dynamic programming techniques are not feasible provided constraints on computing resources. \citep{Chang:2005} demonstrates the specific use of the bandit algorithm, especially its ability to converge to the optimal policy of an MDP. Much research, such as \citep{Coquelin:2007} and \citep{Kocsis:2006}, demonstrates the effective usage of MCTS to obtain $\pi(S, a)$. The primary advantage of MCTS over dynamic programming solutions of the MDP is that, when implemented correctly, the number of iterations through the state space is drastically reduced. Nevertheless, the use of DP or MCTS is not mutually exclusive, as the combination of both algorithms can yield specific advantages and/or disadvantages \citep{Feldman:2014}. The key notion of MCTS is the ability to guide the MC search down the tree efficiently, which borrows itself from the bandit strategy UCB1 \citep{Auer:2002}.

\begin{equation} \label{eq:mcts-uct}
UCT = E(S') + C \sqrt{\frac{2 \ln {n}}{n'}}
\end{equation}

\begin{figure}[!h] 
  \includegraphics[width=127mm]{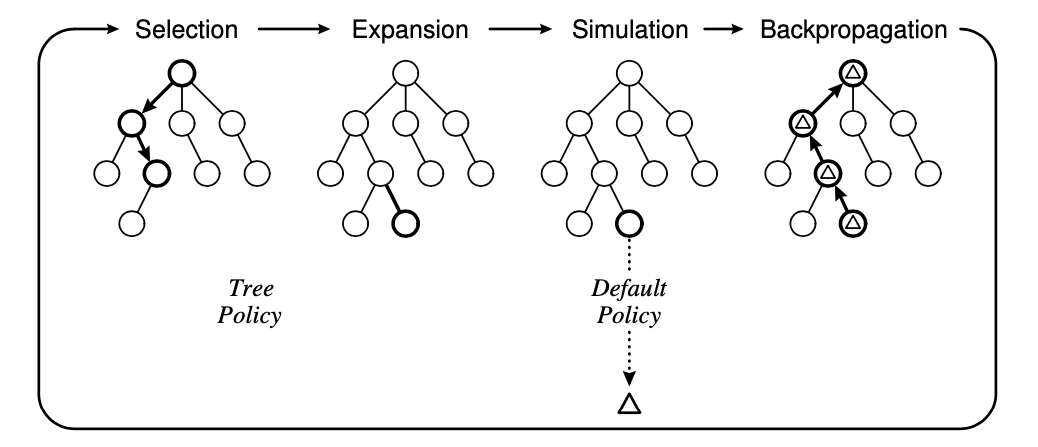}
  \centering
  \caption{Outline of MCTS - from \citep{Browne:2012}.}
  \label{fig:mcts-diagram}
\end{figure}

MCTS primarily makes use of a deterministic selection of actions and resulting outcomes to estimate the reward function of the system. MCTS is a tree search adaptation of the UCB1 Multi-Armed Bandit Strategy \citep{Auer:2002}. When performing one tree traversal in MCTS, a series of actions is randomly played. This tree search is not entirely random as it is guided by the UCB1 illustrated in Eq. (\ref{eq:mcts-uct}). The MCTS algorithm is distinctly divided into 4-phases, \textit{Selection}, \textit{Expansion}, \textit{Simulation}, and \textit{Backpropagation}, which are clearly illustrated in Fig. \ref{fig:mcts-diagram}. In \textit{Selection}, a policy deterministically selects which action to play, based on previously expanded states. This selection is typically guided by the UCT measure, from Eq. (\ref{eq:mcts-uct}). In the \textit{Expansion} phase, states that are unexplored, represented by a leaf node, are added to the search tree one at time. Subsequently, in the \textit{Simulation} phase, a simulation is stochastically played out. Finally \textit{Backpropagation} propagates the final reward of either a terminal state, or a node at an arbitrary depth limit, back to the root node. This 4-phase process is repeated until a maximum number of iterations or a convergence criteria is established. 

\section{Software Framework} \label{sec:software}

\textit{mctreesearch4j} is designed follow three key design principles. Firstly, we design for adaptability. Adaptability is defined as the ability for MDP domain to be easily integrated into the \textit{mctreesearch4j} framework with ease via class abstractions. Our implementation seeks to simplify the adoption of MCTS solutions for a variety of domains. Secondly, we design a software that is fully compatible with the Java Virtual Machine (JVM), for the reasons to be discussed in Section \ref{sec:jvm}. And lastly, we design to achieve a high degree of extensibility within the framework. Extensibility is the defined as the ability for key mechanisms to be reused, redefined, and enhanced, without sacrificing interoperability.

\subsection{JVM Library} \label{sec:jvm}

Open source implementations of MCTS exist, but have not gained widespread adoption. \citep{Cowling:2012} presents an implementation in Python, where multiple game engines can be plugged into the MCTS solver. \citep{Lucas:2018} presents a simple implementation written in Java, written with some categorization of the key MCTS mechanisms outlined in Fig. \ref{fig:mcts-diagram}. However, both implementations do not provide easy access to heuristics, nor do they implement extensible and modular state action space abstractions. \cite{Chatzilygeroudis:2016} presents a low-level implementation of MCTS in C++. Yet because of its low-level implementation, it is not easy to interface a custom MDP definitions with the MCTS solver, hindering its adaptability across domains. Some MCTS implementations are available in Python, a widely adopted scripting language popular in Machine Learning, but suffer performance issues from utilizing an interpreter, causing computing overhead at runtime. Some workarounds do exist, such as dispatching to more performant libraries written in compiled code, but this is additional complexity we desire to avoid. Therefore, compiled languages such as Java or C are preferred for MCTS implementation primarily because of its more efficient resource consumption at runtime compared to native Python. Thus in the current landscape, there is a lack of open source MCTS implementations that provides the full list of features offered by \textit{mctreesearch4j} outlined in Section \ref{sec:software}. 

As of today, JVM language are widely adopted among many open source frameworks\footnote{\url{https://www.tiobe.com/tiobe-index/}}. JVM languages are used in a variety of workloads including web, desktop and Android applications etc. The long history of the Java language also contributes to the JVM extensive knowledge base, language support and backwards compatibility. Furthermore, JVM is known for its wide adoption particularly in the creation of mobile games, many of which require online and/or offline planning solvable via MCTS. Therefore, we chose to develop an MCTS Framework for the JVM ecosystem. 

The Kotlin programming language was chosen for the development of \textit{mctreesearch4j}. Kotlin features a more modern syntax that is succinct and expressive, containing powerful primitives, while simultaneously supporting a wide range of programming paradigms. This language is also popular among many software developers, particularly for the Android mobile operating system, a platform where \textit{mctreesearch4j} may find suitable adoption. The full compatibility of Kotlin with other JVM languages also allows the adoption by other languages within the JVM ecosystem. In addition Kotlin, along with many JVM languages, has strong support for type generics enabling us to build and reuse functions independent of types. It also offers better development experience with compile time type checks without explicit casts.

\subsection{MDP Domain Abstraction} \label{sec:mdp-domain-abstraction}

Attempts to rigorously define MDP interfaces to plug into custom MDP solvers have been attempted in the past. For example, RDDL \citep{Sanner:2010} provides an interface for a formal MDP definition, but it is rather complex for straightforward software implementations, and also adds additional overhead by introducing a new programming language specifically for the definition of MDP's. We simplify the formal MDP definition by abstracting the MDP into its base characteristics outlined in Listing \ref{code-listing:mdp-abstraction}. 

The main abstraction that is used to define an MDP problem is the abstract class defined in Listing \ref{code-listing:mdp-abstraction}, using generic type variables. Each of the methods correspond to specific behaviour of a discrete MDP. In \textit{mctreesearch4j} we use generic types \mintinline{kotlin}{StateType} and \mintinline{kotlin}{ActionType} to represent the MDP states and actions respectively. This abstract class has five members that must be implemented. These abstract class methods define the functionality of an MDP. The MDP abstraction will be used by core MCTS solvers to compute the optimal policy. The MDP interface can be written in any JVM language, we use Kotlin and Scala for this paper, with the Scala implementation from \citep{Liu:2021-connect4}.

\begin{figure}[!h] 
  \includegraphics[width=155mm]{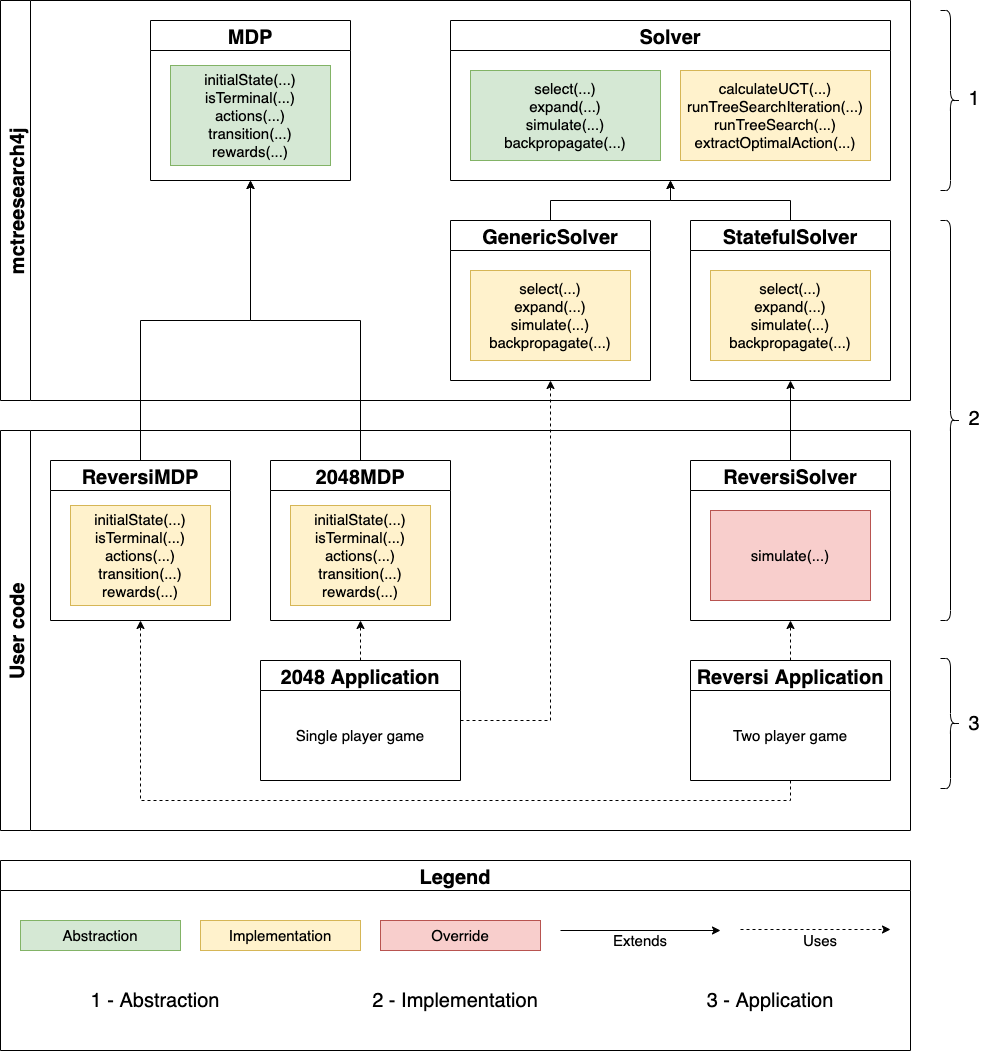}
  \centering
  \caption{Inheritance and class structure of \textit{mctreesearch4j}.}
  \label{fig:software-design}
\end{figure}

\textit{mctreesearch4j} is designed to easily adapt to a variety of MDP definitions without complex mathematical constructs. The computational behaviour of \textit{mctreesearch4j} is defined separately by user code outside of the core library, for example via a game controller, see Fig. \ref{fig:software-design}. This allows programmers to define complex flexible systems, integrating seamlessly into turn-based game applications, including single and multi agent turn-based games such as 2048 and Reversi (Othello), respectively. In the user code layer of the framework, the programmer can define custom solvers, where unique heuristics can be implemented to add domain specific knowledge to improve the performance of the base MCTS solver. Thus the adaptability of the solver to a wide variety of MDP domains is one key design principle of \textit{mctreesearch4j}, see Section \ref{sec:benchmark} for an overview of adapted domains.

\textit{mctreesearch4j} is able to use the MDP definitions to execute a tree search over the state action space. The MDP definitions used to interface with \textit{mctreesearch4j} should optimize its save and load functionalities, as it can impact MCTS search performance if poorly constructed. Nevertheless, defining the MDP space programmatically gives the strong advantage of having the ability to define complex MDP's without the need for rigorous mathematical definitions. In our implementation, we aim to define a clear common declarative interface to provide abstractions for the solver and the MDP representation. The MDP abstraction ensures that the user defines the MDP scenario consistently such that it can be queried by the default solver provided in the library. 

\begin{listing}
\begin{minted}[mathescape, linenos]{kotlin}

abstract class MDP<StateType, ActionType> {
    abstract fun transition(StateType, ActionType) : StateType
    abstract fun reward(StateType, ActionType?, StateType) : Double
    abstract fun initialState() : StateType
    abstract fun isTerminal(StateType) : Boolean
    abstract fun actions(StateType) : Collection<ActionType>
}
\end{minted} 
\caption{Definition of the MDP abstract class.} 
\label{code-listing:mdp-abstraction}
\end{listing}

\subsection{MCTS Solver Design} \label{sec:solver-abs}

\textit{mctreesearch4j} also implements the core Monte Carlo Tree Search algorithm in an extensible paradigm, maximizing the flexibility of the implementation by following Object Oriented Programming (OOP) principles. This is useful because it is possible to run MCTS with customized key mechanisms, and it is highly efficient if the programmer is able to reuse, and/or redefine complete or partial segments of the key mechanisms. In Listing \ref{code-listing:key-mechanisms} we define such key mechanisms, all of which are inherited, and can be redefined or enhanced. This is important both for the efficiency of development, as well as the interoperability of the key MCTS mechanisms.

Listing \ref{code-listing:key-mechanisms} defines four abstract methods which are used to implement MCTS as illustrated in Fig. \ref{fig:mcts-diagram}. The 4 phases, \textit{Selection}, \textit{Expansion}, \textit{Simulation}, and \textit{Backpropagation} defined in lines \ref{code:select} to \ref{code:backpropagate} serve as the key mechanisms to run \textit{mctreesearch4j} and the class method \mintinline{kotlin}{fun runTreeSearch()} is a predefined method from the base \mintinline{kotlin}{class Solver()} responsible for executing the key MCTS solver mechanisms defined in Lines \ref{code:select} to \ref{code:backpropagate}. These open functions, which can be overridden in subclasses, provide the flexibility to define the implementation of the solver that is appropriate for the MDP domain. Furthermore, we introduce the \mintinline{kotlin}{NodeType} which is an abstract representation of the search space. It denotes how the state action space is represented in stages of the MCTS search, and must be explicitly defined by a subclass.

\begin{listing}
\begin{minted}[mathescape, linenos, escapeinside=!!]{kotlin}

abstract class Solver<ActionType, NodeType>(
        protected val verbose: Boolean,
        protected val C: Double // Exploration Constant
        ) where NodeType : Node<ActionType, NodeType> {
    protected abstract var root: NodeType
    abstract fun select(node: NodeType) : NodeType !$\label{code:select}$!
    abstract fun expand(node: NodeType) : NodeType
    abstract fun simulate(node: NodeType) : Double
    abstract fun backpropagate(node: NodeType, reward: Double) !$\label{code:backpropagate}$!
    open fun runTreeSearch(iterations: Int) {/* Predefined method */} 
}

\end{minted} 
\caption{ Key Mechanisms of an MDP Solver } 
\label{code-listing:key-mechanisms}
\end{listing}

\textit{mctreesearch4j} provides a default implementation known as \mintinline{kotlin}{class GenericSolver}, and an alternate \mintinline{kotlin}{class StatefulSolver}. The abstract \mintinline{kotlin}{class Solver} serves as the base class for both versions, and defines a set of functionalities that all solver implementations must provide as well as a set of extensible utilities. Similar to the MDP abstraction, the solver uses a set of type parameters to provide strongly typed methods that unify and simplify the MCTS implementation. The programmer is free to select the data type or data structure that best defines how states and actions are represented in their MDP domain. Thus we can infer that, Generic and Stateful solvers have different representations of the \mintinline{kotlin}{NodeType}. 

The differentiation lies in their respective memory utilization of abstract node types to track states during MCTS iterations. The default \mintinline{kotlin}{class GenericSolver} provides a leaner implementation, where actions are tracked and no explicit states are stored permanently. The states tracked with \mintinline{kotlin}{class GenericSolver} are dynamic and the MDP transitions must be rerun when traversing the search tree during selection in order to infer the state. The \mintinline{kotlin}{class StatefulSolver} keeps an explicit record of the visited states, and additional information, such as terminality and availability of downstream actions. The extra overhead of storing the state explicitly in the MCTS node, allows the algorithm to optimize its search using information from previously visited states. This is particularly useful for deterministic games, where a re-computation of the MDP transition is not necessary to determine the state of the agent after a particular taking a specific action. This differentiation results in different implementations of the \textit{Selection} step, while maintaining identical implementations of \textit{Expansion}, \textit{Simulate} and \textit{Backpropagation}. Both implementation iterates the algorithm in the most common pattern of \textit{Selection}, \textit{Expansion}, \textit{Simulation}, \textit{Backpropagation} until a certain number of \mintinline{kotlin}{iterations}.

\subsubsection{Generic Solver Design} \label{sec:generic-solver-design}

The Generic solver provides an MCTS solution where the states are not explicitly stored in the search tree, but instead inferred from the actions taken by the agent. Generic solvers are more memory efficient, and are preferred for scenarios of high state State-Action-State (SAS) branching factor. We define SAS branching factor as the average number of possible states that the agent can arrive at, given any action from any state. For stochastic domains, any action can lead to a multitude of states, so storing states explicitly in the search tree can be infeasible. In this case the Generic solver is preferred as it obviates the need to track states explicitly. A downside of the Generic solver is that additional computation is required to infer the state of the agent during the \textit{Selection} phase of MCTS, as accomplished via the method on Line \ref{code:generic-simulateActions} of Listing  \ref{code-listing:generic-selection}. The Generic solver can also be extended to Partially Observable MDP's (POMDP's), where we abstract away the state space altogether, and solve over the estimation of state space as inferred by the action reward history, this is also known as the \textit{Belief MDP} approach. In the Belief MDP, a POMDP is simplified into an MDP, by ignoring the potential discrepancy between the observation and true state, treating the current observation as the true state. This can create a tractable MDP solution at the cost of strict accuracy, and is in principle, fully programmable into \textit{mctreesearch4j} via the Generic solver.

\begin{listing}
\begin{minted}[mathescape, linenos, escapeinside=!!]{kotlin}

fun select(node: ActionNode): ActionNode {
    if (node.getChildren().isEmpty()) {return node}
    var currentNode = node
    simulateActions(node) !$\label{code:generic-simulateActions}$!
    while (true) {
        if (mdp.isTerminal(currentNode.state)) {return currentNode}
        val currentChildren = currentNode.getChildren()
        if (currentNode.isNotFullyExplored()) {return currentNode}
        currentNode = currentChildren.maxByOrNull{a -> calculateUCT(a)}
        simulateActions(currentNode)
    }
}

\end{minted} 
\caption{Generic Solver Selection.} 
\label{code-listing:generic-selection}
\end{listing}

The most significant difference is the trade-off between memory and runtime. The Generic solver does not track the possible states for each node of the search tree and therefore reduces the memory consumption. The key benefit is the ability to track the state space with actions alone. A disadvantage, is that the agent must replay the game engine to approximate the state at any time, which is where a computational bottleneck can arise. 

\subsubsection{Stateful Solver Design} \label{sec:stateful-solver-design}

We provide an alternate solver design targeted for MDP domains with low SAS branching factors. In \mintinline{kotlin}{StatefulSolver} the nodes of the solver explicitly track the state, the metadata contained within the state. An \mintinline{kotlin}{NodeType} is used for Stateful search trees where the node explicitly contains both the state of the agent, and the inducing action that led to said state. The Stateful solver is computationally efficient when the SAS branching factor is feasible, as it is able to store and look up the current state of the agent and continue MCTS iterations from the recorded state. Stateful solvers are generally preferred for low state space MDP's, where the SAS branching factor, defined earlier in Section \ref{sec:generic-solver-design}, is relatively manageable, such as in games where low SAS branching factors are exhibited. However, this optimization will incur additional memory usage. 

Listing \ref{code-listing:stateful-selection-expansion} illustrates the \textit{Selection} and \textit{Expansion} algorithm for Stateful MCTS. The Stateful solver keeps track of all possible states at each node illustrated in Lines \ref{code:stateful-sa1} to \ref{code:stateful-sa2}. In the \textit{Selection} phase, the Stateful solver iterates until a leaf node is reached. 

\begin{listing}
\begin{minted}[mathescape, linenos, escapeinside=!!]{kotlin}

fun select(node: StateNode): StateNode { 
    if (mdp.isTerminal(node.state)) {return node} !$\label{code:stateful-sa1}$!
    if (node.validActions.size > node.exploredActions.size) {return node}
    var bestAction = exploredActions.getActionUCT(node, node.getChildren())
    val newState = mdp.transition(node.state, bestAction)
    val actionState = node.getChildren(bestAction)
        .firstOrNull { s -> s.state == newState }
        ?: return createNode(node, bestAction, newState)
    return select(actionState)
} !$\label{code:stateful-sa2}$!


\end{minted} 
\caption{Stateful Selection.} 
\label{code-listing:stateful-selection-expansion}
\end{listing}

\subsubsection{Expansion, Simulation and Backpropagation}

The \textit{Expansion}, \textit{Simulation}, and \textit{Backpropagation} steps for Generic and Stateful solvers remain virtually identical, notwithstanding specification of a different \mintinline{kotlin}{NodeType} for each solver, also resulting minor nuances in instantiating said nodes in the \text{Expansion} phase. The base \mintinline{kotlin}{NodeType}, which is used by both the Generic and Stateful Solvers, store key metadata such as the inducing action, expected reward, and number of visits. The \mintinline{kotlin}{fun expand()} method randomly selects an unexplored node, and adds it to the search tree. This is done using the \mintinline{kotlin}{fun createNode()} method, which differs between Generic and Stateful solvers. 

In the Generic solver, no actual MDP state is required to create a new node of type \mintinline{kotlin}{ActionNode}. The current state of the agent is computed based on the sequence of actions taken utilizing the specified \mintinline{kotlin}{mdp} object, defined in Listing \ref{code-listing:mdp-abstraction}. The Stateful solver, on the other hand, explicitly requires the tracking of states when creating a \mintinline{kotlin}{StateNode} via the \mintinline{kotlin}{fun createNode()} method. The \mintinline{kotlin}{fun simulate()} method runs a purely randomized exploration of the MDP state-action space from a specific state, until a terminal state or the simulation depth is reached. The \mintinline{kotlin}{fun backpropagate()} method is used to update the reward values for the node where the simulation was run as well as all of its ancestors. 

\begin{listing}
\begin{minted}[mathescape, linenos]{kotlin}

fun expand(node: StateNode): StateNode {
    if (node.isTerminal) {return node}
    val unexploredActions = node.getUnexploredActions()
    val actionTaken = unexploredActions.random()
    return createNode(node, actionTaken)
}

fun simulate(node: NodeType): Double {
    if (mdp.isTerminal(node.state)) {
        return mdp.reward(node.parent?.state, node.inducingAction, node.state)
    }
    var depth = 0
    var currentState = node.state
    var discount = rewardDiscountFactor
    while(true) {
        val validActions = mdp.actions(currentState)
        val randomAction = validActions.random()
        val newState = mdp.transition(currentState, randomAction)
        if (mdp.isTerminal(newState)) {
            return mdp.reward(currentState, randomAction, newState) * discount
        }
        currentState = newState
        depth++
        discount *= rewardDiscountFactor
        if (depth > simulationDepthLimit) {
            return mdp.reward(currentState, randomAction, newState) * discount
        }
    }
}

fun backpropagate(node: NodeType, reward: Double) {
    var currentNode = node
    var currentReward = reward
    while (true) {
        currentStateNode.maxReward = max(currentReward, currentNode.maxReward)
        currentStateNode.reward += currentReward
        currentStateNode.n++
        currentStateNode = currentStateNode.parent ?: break
        currentReward *= rewardDiscountFactor
    }
}
\end{minted} 
\caption{Expansion, Simulation and Backpropagation pseudo-code.} 
\label{code-listing:sim-backprop}
\end{listing}

\subsubsection{Extensibility and Heuristics} \label{sec:heuristics}

\begin{figure}[!h] 
    \centering
  \includegraphics[width=120mm]{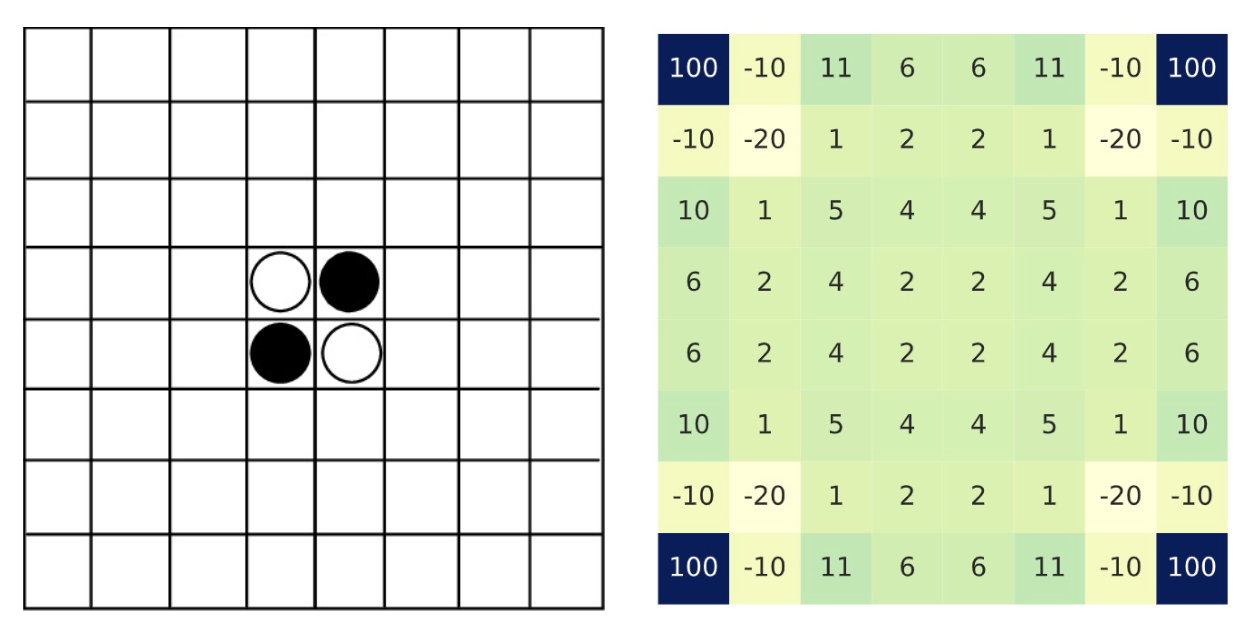}
  \caption{Reversi \mintinline{kotlin}{heuristicWeight} lookup table.}
  \label{fig:reversi-heu}
\end{figure}

Though the default MCTS implementation described in Section \ref{sec:mcts_intro} works well in many scenarios, there are situations where knowledge about specific problem domains can be applied to improve the MCTS performance. Improvements to MCTS, such as heuristics driven simulation, exploit domain knowledge to improve solver performance performance. We demonstrate that a Reversi AI that uses heuristics derived from \citep{Guenther:2004} is able to outperform the basic MCTS implementation. These heuristics are programmed via extensibility points in the \textit{mctreesearch4j} solver implementation, where the key mechanisms can be altered or augmented. Our heuristic, illustrated in Fig. \ref{fig:reversi-heu}, introduces the \mintinline{kotlin}{heuristicWeight} array, a 2D array storing domain specific ratings of every position on a Reversi board representing the desirability of that position on the board. The negative numbers represent a likely loss and positive numbers representing a likely win, again as represented in Fig. \ref{fig:reversi-heu}. This value is taken into consideration when traversing down the simulation tree. The \mintinline{kotlin}{heuristicWeight} array adjusts the propensity to explore any position for both agents based on the heurisitc's belief about the desirability of the position. The effectiveness of this heuristic is later presented in Section \ref{sec:adversarial-bench}. To alter the MCTS simulation phase we override the \mintinline{kotlin}{fun simulate()} method and create a new definition for it. The application of this \mintinline{kotlin}{heuristicWeight} only requires minor alterations to the \mintinline{kotlin}{fun simulate()} method, as illustrated from Lines \ref{code:heu-start} until \ref{code:heu-end} in Listing \ref{code-listing:reversi-heu}.

The default solver behaviour continues running the MCTS iteration until the iteration count is reached. This may not be the best criteria and convergence metrics can be used to better determine when to stop the tree construction. For example, if no significant improvements in the best action's score is made, the algorithm can be terminated early. Additional functionality can also be added to the solver in the tree construction function. For example, if additional metrics need to be measured after each MCTS iteration, \textit{mctreesearch4j} can be overridden to add this functionality. For example, the exploration term, $C$, from Eq. \ref{eq:mcts-uct} can be exported via a class extension that tracks the exploration term.

\begin{listing}
\begin{minted}[mathescape, linenos]{kotlin}

override fun simulate(node: NodeType): Double {
    /*... Original Simulation code ...*/
    while(true) {
        val validActions = mdp.actions(currentState)
        var bestActionScore = Int.MIN_VALUE // Begin heuristic $\label{code:heu-start}$
            var bestActions = mutableListOf<Point>()
            for (action in validActions) {
                val score = heuristicWeight[action.x][action.y]
                if (score > bestActionScore) {
                    bestActionScore = score
                    bestActions = mutableListOf(action)
                }
                if (score == bestActionScore) {bestActions.add(action)}
            }
        val randomAction = bestActions.random() // End heuristic $\label{code:heu-end}$
        val newState = mdp.transition(currentState, randomAction)
        /*... Original Simulation code ...*/
    }
}
\end{minted} 
\caption{Heuristic implementation in Reversi.} 
\label{code-listing:reversi-heu}
\end{listing}

\section{Evaluation \& Function} \label{sec:benchmark}

We evaluate our MCTS implementation on its ability to converge to an accurate solution. In our experiments, the MDP has one or many optimal solutions, and the solver should accurately converge to these solutions. We also measure the performance of the implementation by the average time to extract one optimal action taken from any state, we denote this simply as \textit{runtime per decision} or $\Delta_d$. $\Delta_d$ is an estimate of the time to make a decision averaged over 100 trials. This can vary from computer to computer, but serves as a rough estimate. We demonstrate that \textit{mctreesearch4j} is adaptable across multiple domains, in addition to the fact that $\Delta_d$ can be kept at manageable levels for operation on a common laptop computer. The exact hardware specifications are outlined in Appendix \ref{app:sim-hardware},

For adversarial search domains featuring two agents, the terminal state of victory or loss results in a score of +1 or -1 respectively, from the perspective of the agent when competing against the opponent. Later in Section \ref{sec:adversarial-bench} we demonstrate a adversarial setup where 2 MCTS solvers with novel characteristics, play against each other. 

We examine the performance of the MCTS solver across five distinct MDP domains. These domains are GridWorld, a probabilistic single agent MDP illustrated in Fig. \ref{fig:grid}, the game of 2048 analyzed in \cite{Kohler:2019}, Push Your Luck (PYL), a single agent MDP defined in \cite{Geisser:2018}, the game of Connect 4, a deterministic multi-agent game, and the game of Reversi, a deterministic adversarial multi-agent game.

\subsection{Solver Runtime per Domain} \label{sec:runtime}

\begin{table}[h!]
    \centering
    \begin{tabular}{||c c c c c c c||} 
    \hline
    Domain & No. Agents & Stochastic & Avg. Br. Fac. & Avg. Depth & Solver & $\Delta_d$  \\ [0.5ex] 
    \hline\hline
    GridWorld & 1 & yes & 4 & 5  & Gen. & 0.1 \\
    PYL & 1 & yes & 2 & $\infty$ & Gen. & 0.3 \\
    2048 & 1 & yes & 4 & 940 & Gen. & 0.5 \\
    Connect4 & 2 & no & 4 & 36 & SF & 0.06 \\
    Reversi & 2 & no & 10 & 58 & SF & 0.12 \\
    \hline
    \end{tabular}
    \caption{MDP domain characteristics, where $\Delta_d$ is in seconds.}
    \label{table:runtime}
\end{table}

Furthermore, we select either Generic or Stateful solvers based on the characteristics of the MDP domain, with the considerations outlined in Sections \ref{sec:stateful-solver-design} and \ref{sec:generic-solver-design}. The max tree depth of the MCTS was set to 1000, this number typically far exceeds the average depth before reaching a terminal state for most MDP domains, with the exception of 2048. The number of iterations set to 500 arbitrarily. The MDP domains, where \textit{mctreesearch4j} was applied, contain one or two agents, and state action transitions can be stochastic or deterministic. We also note the average branching factor of the domain (Avg. Br. Fac.), and average depth (Avg. Depth)\footnote{\mintinline{bash}{https://en.wikipedia.org/wiki/Game_complexity} illustrates Avg. Br. Fac. and Avg. Depth.} of the search tree until a terminal state is reached. For some games, such as GridWorld, average depth is heavily determined by the state configuration, we refer to the configuration set out in Fig \ref{fig:grid}. We record the average runtime to compute the immediate next optimal action from an arbitrary state in the state space of the MDP domain in milliseconds\footnote{The exact specification of the computer used to run the simulation is contained in Appendix \ref{app:sim-hardware}.}.

\subsection{Solver Convergence}  

To demonstrate the accuracy of \textit{mctreesearch4j}, we examine the performance of the solver with respect to convergence of rewards, convergence of exploration terms, and visiting of the optimal state subspace more than non-optimal state subspace with increasing iterations. Variables such as the exploration constant $C$, as denoted in Eq. (\ref{eq:mcts-uct}), affect the overall exploration of \textit{mctreesearch4j}. If $C$ is too high, there can be over-exploration causing the UCT function to swing between optimal actions and non-optimal action selection. Over-exploration affects the smoothness of the exploration term convergence \citep{James:2017}. If $C$ is too low, then it is not able to sufficiently explore all states to find an optimal action. 

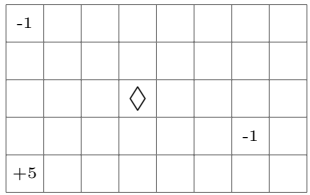
\begin{figure}[!h]
\centering
\begin{tikzpicture}
\draw[step=0.5cm,color=gray] (0,0) grid (4,2.5);

\node at (0.25 + 0.5*0 , 0.25 + 0.5*0) {\tiny +5};
\node at (0.25 + 0.5*3 , 0.25 + 0.5*2) {$\Diamond$};
\node at (0.25 + 0.5*0 , 0.25 + 0.5*4) {\tiny -1};
\node at (0.25 + 0.5*6 , 0.25 + 0.5*1) {\tiny -1};

\end{tikzpicture}
\caption{Illustration of the Gridworld MDP domain. An agent, is represented by $\Diamond$, lies in a domain with terminal state reward +5, and terminal reward negative state -1.}
\label{fig:grid}
\end{figure}

We provide a simple example of convergence evaluation in the domain of Gridworld. The grid displayed in Fig. \ref{fig:grid}, represents the state space, and each cell demarcates the state which corresponds to a reward value (unfilled cell indicates a reward of 0). In our example any non-zero state represents a terminal state. The agent represented by $\Diamond$, has 4 possible actions from its starting point. In the specific configuration shown by Fig. \ref{fig:grid}, the MCTS solver should ultimately select the actions ← or ↓, which are optimal, and \textit{mctreesearch4j} should accurately converge to the appropriate values.

\begin{figure}[H] 
  \includegraphics[width=\textwidth]{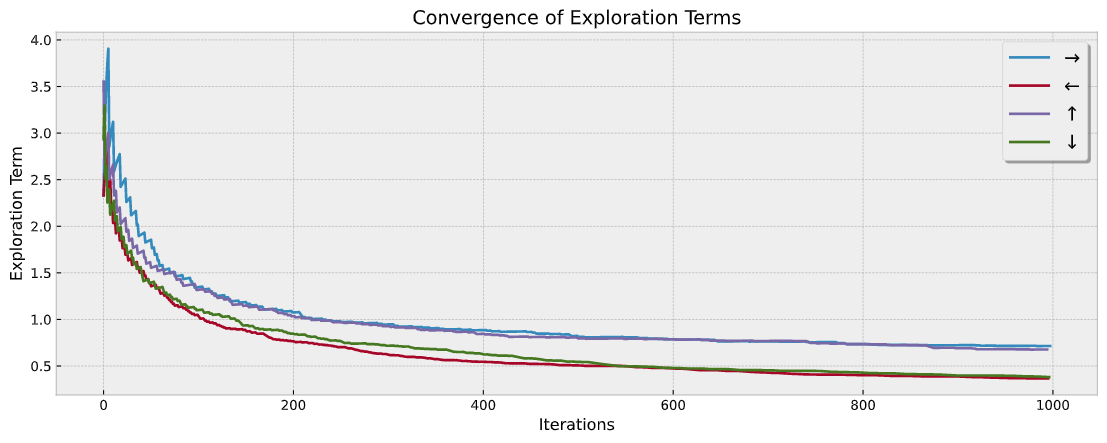}
  \caption{Convergence of exploration terms}
  \label{fig:gw-exp-terms}
\end{figure} 

\begin{figure}[H] 
  \includegraphics[width=\textwidth]{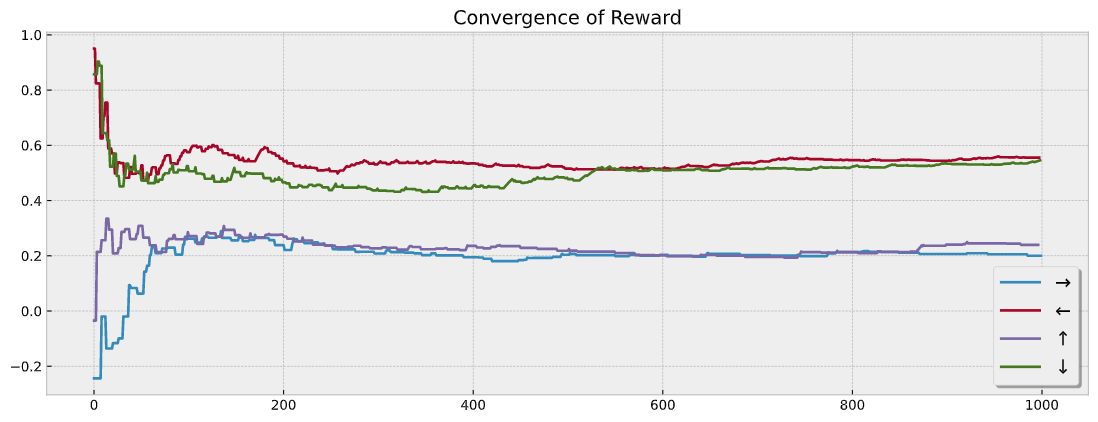}
  \caption{Convergence of rewards}
  \label{fig:gw-rewards}
\end{figure}

\begin{figure}[H] 
  \includegraphics[width=\textwidth]{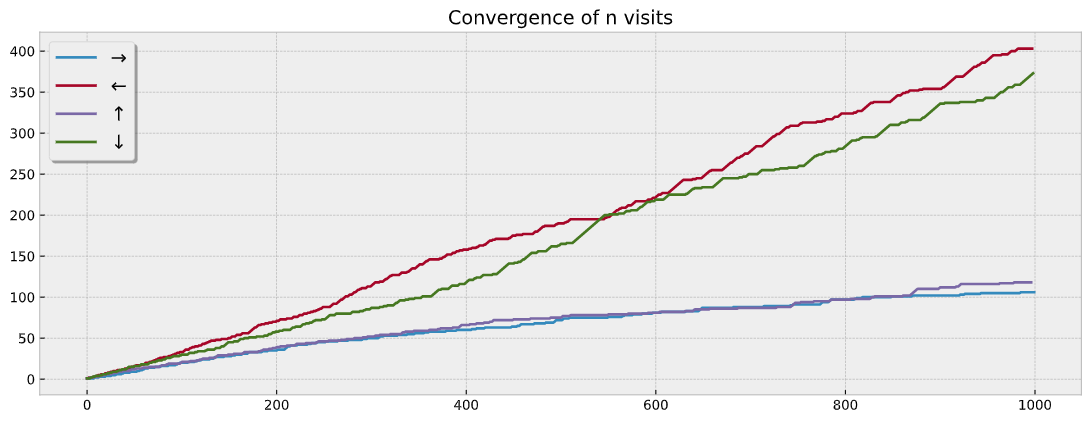}
  \caption{Convergence of visits}
  \label{fig:gw-visits}
\end{figure}

Our MCTS solver\footnote{The figures \ref{fig:gw-exp-terms} to \ref{fig:gw-visits} represent solutions of the Stateful solver, nevertheless convergence plots using Generic solver yields similar behaviour.}, with no additional heuristics, is capable of converging to the correct solution space. Fig. \ref{fig:gw-exp-terms} displays a convergence of the exploration term $C \sqrt{2 \log{n} / n'}$, which decreases as $n \rightarrow \infty$, and $n' \leq n$. Provided this, the correct optimal policy will yield smaller exploration terms for optimal versus non-optimal solutions because the number of visits to the non-optimal nodes, $n'$, will be generally smaller than the optimal nodes.

Naturally, the reward value associated with each optimal state induced by the optimal policy action at $t + 1$ should yield higher backpropagated rewards. In Fig. \ref{fig:gw-rewards} we see that the two optimal actions produce higher reward values than non-optimal solutions.

When the MCTS solver is accurately selecting the optimal solutions, it will continue to cause the agent to explore in the optimal subset of the state space, and reduce its exploration in the non-optimal subset of the state space as evidenced in Fig. \ref{fig:gw-visits}. The cumulative number of visits corresponding to the optimal policy is proportionally increasing with respect to the number of MCTS iterations. Whereas for non-optimal solutions, the cumulative visits are significantly less because the solver will avoid visiting the non-optimal state subspace. 

\subsection{Adversarial Game Benchmarking} \label{sec:adversarial-bench}

We demonstrate the possibility of heuristic adaptation to improve \textit{mctreesearch4j} in an adversarial game. For this, we create a two-player game in the deterministic multiplayer domain of Reversi, where our AI will play as adversaries against each other. The Reversi heuristic is implemented to enable the search algorithm to compute more realistic estimates of reward, using specific domain knowledge \citep{Guenther:2004}. The intended effect of this heuristic is to enable a more efficient search of the state action space, as presented earlier in Section \ref{sec:heuristics} 

The heuristic augmented \textit{mctreesearch4j} is played against the original \textit{mctreesearch4j} to prove that the adversarial MDP solver is capable of improving via domain knowledge. Table \ref{table:reversi_results} demonstrates that when \textit{mctreesearch4j} is augmented with heuristics it defeats the original \textit{mctreesearch4j} solver up to 75\% of the time, under ceteris paribus conditions. Nevertheless this win rate metric depends on several factors. Firstly, we see that the heuristic solver is only marginally better than the original solver when we limit the maximum simulation depth (Max. Sim. Depth), and the number of simulation iterations (No. Iter). The limitation on these two hyper-parameters limit the MCTS solver's ability to explore the entire state action space to determine accurate rewards, thereby also limiting the effectiveness of the heuristic. The degree of contrast between \textit{mctreesearch4j} with heuristic and without is greatly enhanced when the solver is provided enough simulation iterations and simulation depth to fully take advantage of the additional domain information.

\begin{table}[h!]
    \centering
    \begin{tabular}{||c c c c c c||} 
    \hline
    No. Iter & Max. Sim. Depth & No. Wins & No. Loss & No. Ties & Win Rate \\ [0.5ex] 
    \hline\hline
    
    50 & 40 & 98 & 87 & 15 & 0.49  \\
    500 & 40 & 121 & 75 & 4 & 0.605  \\
    
    50 & 1000 & 111 & 75 & 14 & 0.555  \\
    500 & 1000 & 151 & 46 & 3 & 0.755  \\
    
    \hline
    \end{tabular}
    \caption{Reversi results comparing MCTS solver with heuristics and without. }
    \label{table:reversi_results}
    \end{table}

Via this experiment, we justify that our MCTS solver is capable of being modified to adapt new heuristics which improve the outcomes of the MDP policy for an adversarial game. The heuristic presented here actively biases the random simulation of MCTS to favour actions which induce better outcomes, based on domain knowledge. We also see that as we adjust hyper-parameters of the model, such as simulation depth and number of iterations, the heuristic augmented solver can become more effective. 

\section{Conclusion \& Future Work}

Due to the extensibility and modularity of \textit{mctreesearch4j}, many improvements and experiments could be performed by extending and modifying the capabilities of the base software. \textit{mctreesearch4j} is designed to enable researchers to experiment with various MCTS strategies while standardizing the functionality of the MCTS solver and ensuring reliability, where the experiments are reproducible and the solver is compatible with common JVM runtimes \footnote{To be specific \textit{mctreesearch4j} is fully compatible with Java Runtime Environment (JRE) 8 and 11.}.

\subsection{Future Work} \label{sec:future-work}

We encourage the effort to enhance \textit{mctreesearch4j} further. For example, while the default implementations create a new node and adds it to the node chosen by the \textit{Selection} phase, it is possible to rely on domain knowledge to avoid superfluous game states. For example, in Gridworld, where each unique position is functionally the same regardless of how it arrived to that position, there is no need to add a newly created node to the search tree. Instead, a cache of nodes representing known positions can be used to dramatically reduce the MCTS search space, yielding more accurate results. Furthermore, creating an mapping for more expressive and rigorous MDP defining meta-languages such as RDDL \citep{Sanner:2010} to bind to \textit{mctreesearch4j}, could be an important feature in the near future. This enables complex and rigorous definitions of MDP's to be benchmarked using \textit{mctreesearch4j}, opening many new opportunities for research.

Also, as described in \cite{Baier:2013}, some games can be solved more efficiently if Minimax algorithms can be applied in the \textit{Selection} phase. Instead of the default solver which uses UCT to select leaf nodes, a Minimax algorithm can be used instead to detect shallow wins and losses. This has been shown to work well for adversarial games to avoid traps set by the opposite player. For adversarial games, another common strategy is to improve simulation accuracy by incorporating a Minimax algorithm for the \textit{Simulation} phase. This ensures the rewards computed are more realistic than two players who play randomly. This can significantly improve the MCTS performance. In addition, the \textit{backpropagation} step can be augmented with Minimax concepts to mark known losing positions. This allows subsequent \textit{Selection} phase to ignore parts of the search tree and speed up convergence \citep{Baier:2013}. Finally, instead of running simulated games to compute a reward, it is possible to rely on an external source to predict the reward value of a given state. An example of this is the state-of-the-art Go AI, AlphaGo \cite{Silver:2016}, which uses Deep Learning trained on expert positions and via self-play to compute the expected reward for a given game position. Combining Deep Learning techniques with \textit{mctreesearch4j} to provide external intelligence is an open avenue of research.

The phase order of MCTS defined in the base \mintinline{kotlin}{class Solver()} can also be modified to explore different avenues. Although the most common MCTS phase order of $\textit{Selection} \rightarrow \textit{Expansion} \rightarrow \textit{Simulation} \rightarrow \textit{Backpropagation}$ works well in most scenarios, the MCTS phase order can be altered to $\textit{Selection} \rightarrow \textit{Simulation} \rightarrow \textit{Expansion} \rightarrow \textit{Backpropagation}$ as described in \cite{Winands:2010}. The advantages and/or disadvantages of altering the MCTS phase order is an under explored topic.

\subsection{Summary}

In closing, \textit{mctreesearch4j} presents a framework which enables programmers to adapt an MCTS solver to a variety of MDP domains. This is important because software application was a main focus of \textit{mctreesearch4j}. Furthermore, \textit{mctreesearch4j} is fully compatible with JVM, and this design decision was made due to the excellent support of class structure and generic variable typing in Kotlin, and other JVM languages, as well as support for mobile applications. Yet most importantly, \textit{mctreesearch4j} is modular and extensible, the key mechanism of MCTS are broken down, and the programmer is able inherit class characteristics, redefine and/or re-implement certain sections of the algorithm while maintaining a high degree of MCTS standardization.

\subsection{Software Availability} \label{sec:library-avail}

\begin{itemize}

\item \mintinline{bash}{https://mvnrepository.com/artifact/ca.aqtech/mctreesearch4j} contains the compiled core library for \textit{mctreesearch4j} available via the Maven Central Repository for the JVM ecosystem.

\item \mintinline{bash}{http://mctreesearch4j.aqtech.ca/} contains the source code for implementation of \textit{mctreesearch4j} core library, as well as programs for various domains and game controllers.
\end{itemize}

\section*{Acknowledgements}

We would like to acknowledge the kind support of our reviewers, and Maven Central Repository for repository hosting of our code base.

\appendix
\section{} \label{appendix}

\subsection{Simulation Hardware Specifications}\label{app:sim-hardware}

All of the benchmarks run in Section \ref{sec:benchmark} were performed on an Apple Macbook Pro, containing a 2.6GHz dual-core Intel Core i5 processor with 3MB shared L3 cache with 8GB of 1600MHz DDR3L onboard memory. We expect that a computer with similar hardware capabilities will yield similar results.


\vskip 0.2in
\bibliography{sample}

\end{document}